\documentclass[runningheads]{llncs}
\usepackage{amsmath}
\usepackage{graphicx}
\usepackage{tabularx}
\usepackage{booktabs} 
\usepackage{hyperref}
\begin{document}
	\title{BiO-Net: Learning Recurrent Bi-directional Connections for Encoder-Decoder Architecture}
	
	\author{Tiange Xiang\inst{1} \and
		Chaoyi Zhang\inst{1} \and
		Dongnan Liu\inst{1} \and Yang Song\inst{2} \and Heng Huang\inst{3,4} \and Weidong Cai\inst{1}}
	% index{Xiang, Tiange}
	% index{Zhang, Chaoyi}
	% index{Liu, Dongnan}
	% index{Song, Yang}
	% index{Huang, Heng}
	% index{Cai, Weidong}
	
	\titlerunning{BiO-Net: Learning for Bi-directional Skip Connections}
	\authorrunning{Xiang et al.}
	\institute{School of Computer Science, University of Sydney, Australia \and School of Computer Science and Engineering, University of New South Wales, Australia \and Electrical and Computer Engineering, University of Pittsburgh, USA \and JD Finance America Corporation, Mountain View, CA, USA \\ \email{\{txia7609, dliu5812\}@uni.sydney.edu.au} \\ \email{\{chaoyi.zhang, tom.cai\}@sydney.edu.au} \\ \email{yang.song1@unsw.edu.au} \\ \email{henghuanghh@gmail.com} }

	\maketitle              % typeset the header of the contribution
	\begin{abstract}
		U-Net has become one of the state-of-the-art deep learning-based approaches for modern computer vision tasks such as semantic segmentation, super resolution, image denoising, and inpainting. Previous extensions of U-Net have focused mainly on the modification of its existing building blocks or the development of new functional modules for performance gains. As a result, these variants usually lead to an unneglectable increase in model complexity. To tackle this issue in such U-Net variants, in this paper, we present a novel \textbf{Bi}-directional \textbf{O}-shape network (BiO-Net) that reuses the building blocks in a recurrent manner without introducing any extra parameters. Our proposed bi-directional skip connections can be directly adopted into any encoder-decoder architecture to further enhance its capabilities in various task domains. We evaluated our method on various medical image analysis tasks and the results show that our BiO-Net significantly outperforms the vanilla U-Net as well as other state-of-the-art methods. Our code is available at \url{https://github.com/tiangexiang/BiO-Net}.
		
	\end{abstract}

	\section{Introduction}
	Deep learning based approaches have recently prevailed in assisting medical image analysis, such as whole slide image classification \cite{zhang2018whole}, brain lesion segmentation \cite{zhang2018ms}, and medical image synthesis \cite{hou2019robust}.
	U-Net \cite{ronneberger2015u}, as one of the most popular deep learning based models, has demonstrated its impressive representation capability in numerous medical image computing studies. U-Net introduces skip connections that aggregate the feature representations across multiple semantic scales and helps prevent information loss.\\
	\noindent\textbf{U-Net Variants.} Recent works were proposed to extend the U-Net structure with varying module design and network construction, illustrating its potentials on various visual analysis tasks. V-Net \cite{milletari2016v} applies U-Net on higher dimension voxels and keeps the vanilla internal structures. W-Net \cite{xia2017w} modifies U-Net to tackle unsupervised segmentation problem by concatenating two U-Nets via an autoencoder style model. Compared to U-Net, M-Net \cite{mehta2017m} appends different scales of input features to different levels, thus multi-level visual details can be captured by a series of downsampling and upsampling layers. Recently, U-Net++ \cite{zhou2018unetpp} adopts nested and dense skip connections to represent the fine-grained object details more effectively. Moreover, attention U-Net \cite{oktay2018attention} uses extra branches to adaptively apply attention mechanism on the fusion of skipped and decoded features. However, these proposals may involve additional building blocks, which lead to a greater number of network parameters and thus an increased GPU memory. Unlike above variants, our BiO-Net improves the performance of U-Net via a novel feature reusing mechanism where it builds bi-directional connections between the encoder and decoder to make inference from a recursive manner.\\
	\noindent
	\textbf{Recurrent Convolutional Networks.} Using recurrent convolution to iteratively refine the features extracted at different times has been demonstrated to be feasible and effective for many computer vision problems \cite{han2018image,guo2019dynamic,wang2019recurrent,alom2018nuclei}. Guo \textit{et al}. \cite{guo2019dynamic} proposed to reuse residual blocks in ResNet so that available parameters would be fully utilized and model size could be reduced significantly. Such a mechanism also benefits the evolution of U-Net. As a result, Wang \textit{et al}. \cite{wang2019recurrent} proposed R-U-Net, which recurrently connects multiple paired encoders and decoders of U-Net to enhance its discrimination power for semantic segmentation, though, extra learnable blocks are introduced as a trade-off. BiO-Net distinguishes R-U-Net from its design of backward skip connections, where latent features in every decoding levels are reused, enabling more intermediate information aggregations with gradients preserved among temporal steps. R2U-Net \cite{alom2018nuclei} adopts a similar approach that only recurses the last building block at each level of refinement. By contrast, our method learnt recurrent connections in the existing encoder and decoder rather than recursing the same level blocks without the involvement of refined decoded features.\\
	\noindent
	To this end, we propose a recurrent U-Net with an bi-directional O-Shape inference trajectory (BiO-Net) that maps the decoded features back to the encoder through the backward skip connections, and recurses between the encoder and the decoder. Compared to previous works, our approach achieves a better feature refinement, as multiple encoding and decoding processes are triggered in our BiO-Net. We applied our BiO-Net to perform semantic segmentation on nuclei segmentation task and EM membrane segmentation task and our results show that the proposed BiO-Net outperforms other U-Net variants, including the recurrent counterparts and many state-of-the-art approaches. Experiments on super resolution tasks also demonstrate the significance of our BiO-Net applied to different scenarios.

	%As one of the earliest proposed recurrent convolutional networks, Romera-Paredes \cite{romera2016recurrent} experimented on object instance segmentation tasks, which results in a great success.
	%We will make a direct comparison between our BiO-Net and R2U-Net in the experiment section to show that our design is achievable for better results.
	
	\section{Methods}
	
	\begin{figure}[t]
		\includegraphics[width=\textwidth]{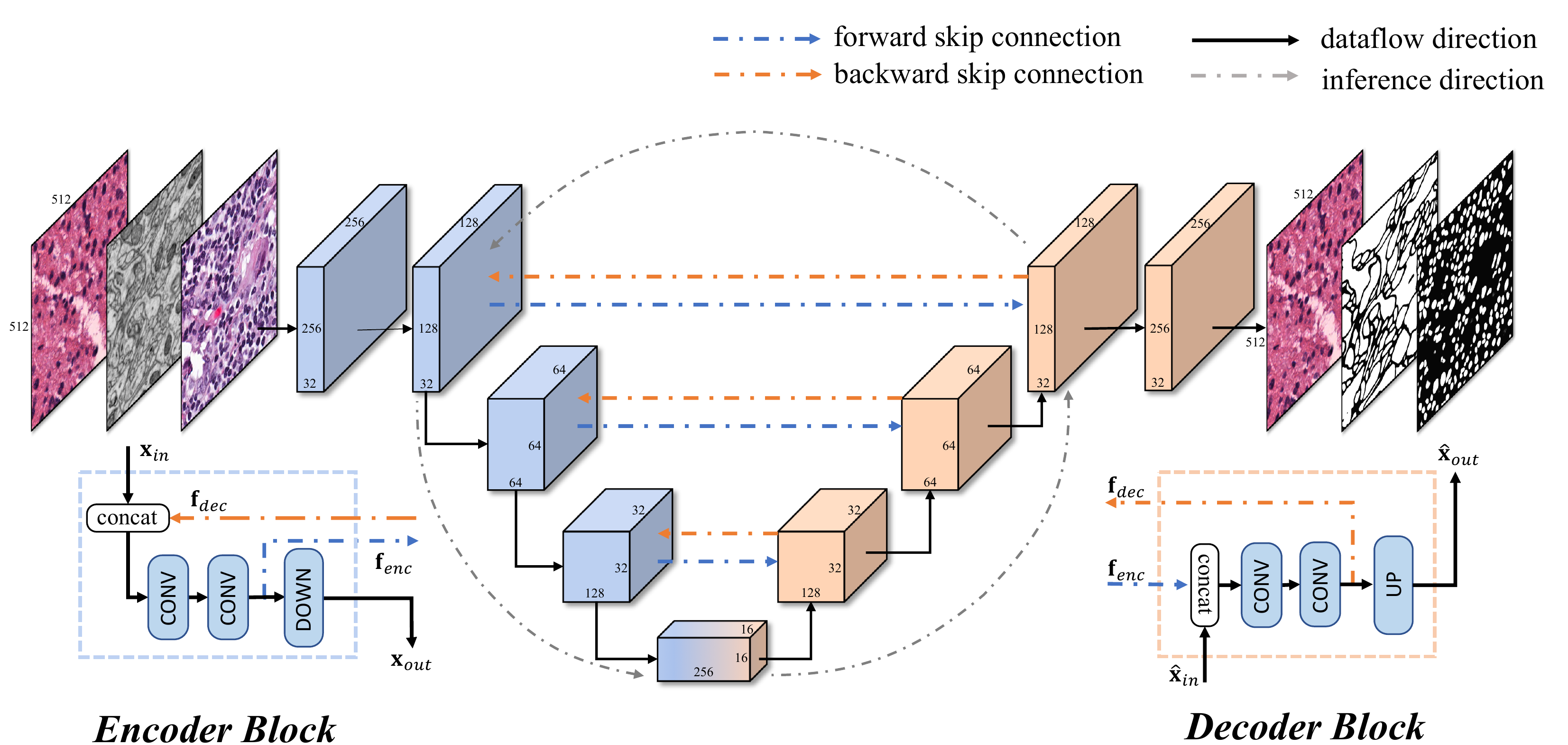}
		\caption{Overview of our BiO-Net architecture. The network inferences recurrently in an O-shape manner. \texttt{CONV} represents the sequence of convolution, non-linearity and batch-norm layers. \texttt{DOWN} stands for downsampling (implemented by a \texttt{CONV} followed by max-pooling), while \texttt{UP} denotes upsampling (archived by transpose convolution).} \label{fig1}
	\end{figure}
	
	% backup: first, with the same number of parameters and model size regardless of the recurrence time \textit{t}, O-Net achieves better performance as \textit{t} increases;
	As shown in Fig \ref{fig1}, BiO-Net adopts the same network architecture as U-Net, without any dependencies on extra functional blocks but with paired bi-directional connections. It achieves better performance as \textit{t} increases with no extra trainable parameters introduced during its unrolling process. Moreover, our method is not restricted to U-Net and can be integrated into other encoder-decoder architectures for various visual analysis tasks.
	
	%allows the cooperation of exterior modules or the modifications of inter-structures, such as residual blocks \cite{he2016deep}, squeeze and excitation blocks \cite{hu2018squeeze} and attention fusion blocks \cite{oktay2018attention} on either forward or backward connections.
	
	\subsection{Recurrent Bi-directional Skip Connections} \label{bc}
	The main uniqueness of our BiO-Net model is the introduction of bi-directional skip connections, which facilitate the encoders to process the semantic features in the decoders and vice versa.\\
	%\noindent
	%\textbf{Forward Skip Connections.} Forward skip connections can preserve low-level features from encoder to decoder and maintain gradients \cite{ronneberger2015u,he2016deep}, by fusing the input tensor $\textbf{x}_l$ to the $l^{th}$ encoder block with the tensor processed by a series of forward convolution operations $\mathcal{F}$, which can be defined as:
	%\textbf{Forward Skip Connections.} Forward skip connections can preserve low-level features from encoder to decoder and maintain gradients \cite{ronneberger2015u,he2016deep}, by fusing the input tensor $\textbf{x}_l$ of the $l^{th}$ encoder block with the decoded features $h_l$ processed by a series of forward convolution operations $\mathcal{F}$, which can be defined as
	%(processed by a series of forward convolution operations $\mathcal{F}$, implicit modelling of operations below)
	\noindent\textbf{Forward Skip Connections.} Forward skip connections linking encoders and decoders at the same level can preserve the encoded low-level visual features $\textbf{f}_{enc}$, with their gradients well-preserved~\cite{ronneberger2015u,he2016deep}. Hence, the $l$-th decoder block could fuse $\textbf{f}_{enc}$ with its input $\hat{\textbf{x}}_{in}$ generated from lower blocks, and propagate them through the decoding convolutions $\texttt{DEC}$ to generate $\textbf{f}_{dec}$, which will be further restored to higher resolutions via $\texttt{UP}$ block. This process can be defined as:
	\begin{equation} \label{eq1}
	\textbf{f}_{dec} = \texttt{DEC}([\textbf{f}_{enc},\ \hat{\textbf{x}}_{in}]),
	%\textbf{f}_{dec} = \texttt{DEC}([\textbf{f}_{enc},\ \mathcal{F}(\textbf{f}_{enc})]),
	\end{equation}  
	where concatenation is employed as our fusion mechanism [$\cdot$]. The index notation ($l$-th) for encoders and decoders is omitted in this paper for simplicity purpose.\\
	\noindent
	\textbf{Backward Skip Connections.} With the help of our novel backward skip connections, which pass the decoded high-level semantic features $\textbf{f}_{dec}$ from the decoders to the encoders, our encoder can now combine $\textbf{f}_{dec}$ with its original input $\textbf{x}_{in}$ produced by previous blocks, and therefore achieves flexible aggregations of low-level visual features and high-level semantic features. %The updated tensor in the $i+1^{th}$ iteration $\textbf{x}^{i+1}$ defined as:
	Similar to the decoding path enhanced by forward skip connections, our encoding process can thus be formulated with its encoding convolutions $\texttt{ENC}$ as:
	\begin{equation} \label{eq2}
	%	\begin{split}
	\textbf{f}_{enc} = \texttt{ENC}([\textbf{f}_{dec}, \ 
	\textbf{x}_{in}]).
	%&= \texttt{concat}(x,\ \texttt{concat}\big(x,\ \mathcal{F}(x)\big),)
	%	\end{split}
	\end{equation}
	The \texttt{DOWN} block feeds $\textbf{f}_{enc}$ to subsequent encoders for deeper feature extraction.
	
	%such that $\textbf{x}^{i+1}_l$ will be the new tensor pass to the same forward convolution blocks at $i+1^{th}$ inference iteration and the $l^{th}$ block. Based on the recurrent skip connections of Eq. \ref{eq1} and Eq. \ref{eq2},  the proposed bi-directional skip connections assist the refinement of feature maps at each resolution iteratively.
	
	\noindent\textbf{Recursive Inferences.} The above bi-directional skip connections create an O-shaped inference route for encoder-decoder architectures. Noticeably, this O-shaped inference route can be recursed multiple times to receive immediate performance gains, and more importantly, this recursive propagation policy would not introduce any extra trainable parameters. Hence, the outputs of encoders and decoders equipped with our proposed O-shaped connections can be demonstrated as follows, in terms of their current inference iteration $i$:
	\begin{equation}
	\begin{split}
	\textbf{x}_{out}^i &= \texttt{DOWN}(\texttt{ENC}([\texttt{DEC}([\textbf{f}_{enc}^{i-1}, \hat{\textbf{x}}_{in}^{i-1}]), \textbf{x}_{in}^i])),\\
	\hat{\textbf{x}}_{out}^i &= \texttt{UP}(\texttt{DEC}([\texttt{ENC}([\textbf{f}_{dec}^i, \ 
	\textbf{x}_{in}^i]),\hat{\textbf{x}}_{in}^i])).
	%\textbf{x}_l^{i} &=  [\mathcal{F}\big(\ [\mathcal{F}(\textbf{x}_l^{i-1} ),\textbf{f}_{l}^{(dec)i-1}] \big), \textbf{x}_{l-1}^{i}].
	\end{split}
	\end{equation}
	%where $\textbf{f}_{dec}^{  i} =  \texttt{DEC}([\textbf{f}_{enc}^i, \hat{\textbf{x}}_{in}^{  i}])$ and $\textbf{f}_{enc}^{i} =  \texttt{ENC}([\textbf{f}_{dec}^i, \textbf{x}_{in}^i])$.
	% which directly defines $\textbf{x}_{out} = \texttt{ENC}(\textbf{x}_{in})$, 
	Compared to the vanilla U-Net, our BiO-Net takes both encoded and decoded features into consideration and performs refinement according to features from the previous iterations. 
	\subsection{BiO-Net Architecture} \label{arch}
	
	We use the plain convolutional layers, batch normalization \cite{ioffe2015batch} layers, and ReLU \cite{nair2010rectified} layers in the network architecture. No batch normalization layer is reused.
	
	\noindent The input image is first fed into a sequence of three convolutional blocks to extract low-level features. Note that there is no backward skip connection attached to the first stage block and, hence, the parameters in the first stage block will not be reused when recursing. The extracted features are then sent to a cascade of encode blocks that utilize max pooling for feature downsampling. During the encoding stage, the parameters are reused and the blocks 
	are recursed through the paired forward and backward connections as shown in Fig. \ref{fig1}. After the encoding phase, an intermediate stage of which contains convolutional blocks that are used to further refine the encoded features. Then, the features are subsequently passed into a series of decode blocks that recover encoded details using convolutional transpose operations. During the decoding stage, our proposed backward skip connections preserve retrieved features by concatenating them with the features from the same level encoders as depicted in Sec. \ref{bc}. The recursion begins with the output generated from the last convolutional block in the decoding stage. After recursing the encoding and decoding stages, the updated output will be fed into the last stage block corresponding to the first stage blocks. Similarly, the last stage blocks will not be involved in the recurrence.
	
\begin{figure}[t]
		\includegraphics[width=\textwidth]{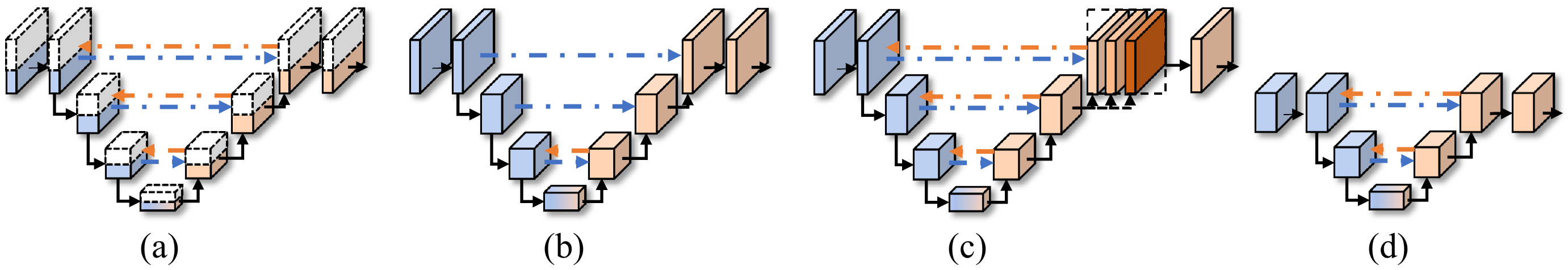}
		\caption{Visualizations of setups in the ablation experiments. (a) BiO-Net with $\times$0.5. (b) BiO-Net with $w$ = 2. (c) BiO-Net with \textit{t} = 3 and \texttt{INT}. (d) BiO-Net with $l$ = 3.} \label{fig2}
	\end{figure}

	\noindent We define several symbols for better indication: `\textit{t}' represents the total recurrence time; `$\times$\textit{n}' represents the expansion multiplier times to all hidden output channel numbers; `$w$' represents the number of backward skip connections used from the deepest encoding level; `\texttt{INT}' represents stacking decoded features from each iteration and feeding them into the last stage block as a whole and `$l$' represents the most encoding depth. Details can be seen in Fig. \ref{fig2}.
	
	\section{Experiments}
	
	\noindent
	\textbf{Datasets.} Our method was evaluated on three common digital pathology image analysis tasks: nuclei segmentation, EM membrane segmentation, and image super resolution on a total of four different datasets. Two publicly available datasets, namely MoNuSeg \cite{kumar2017dataset} and TNBC \cite{naylor2018segmentation}, were selected to evaluate our method for nuclei semantic segmentation. The MoNuSeg dataset consists of a 30-image training set and a 14-image testing set, with images of size $1000^2$ sampled from different whole slide images of multiple organs. We extract $512^{2}$ patches from 4 corners of each image, which enlarges the dataset by 4 times. TNBC is comprised of 50 histopathology images of size $512^2$ without any specific testing set. Both datasets include pixel-level annotation for the nuclei semantic segmentation problem. The second task we evaluated is EM membrane segmentation, where the piriform cortex EM dataset of the mice collected from \cite{lee2015recursive}, which contains four stacks of EM images with the slice image sizes of $255^2$, $512^2$, $512^2$, and $256^2$, respectively. Image super resolution is the last task we evaluated our method on, the dataset was constructed from a whole slide image collected by MICCAI15 CBTC. We sampled 2900 patches of size $512^2$ at $40\times$ magnification level, with the 9 : 1 split ratio for the training and testing set, respectively.\\
	\noindent
	\textbf{Implementation Details.}  \label{imp}
	We used Adam \cite{kingma:adam} optimizer with an initial learning rate of 0.01 and a decay rate of 0.00003 to minimize cross entropy loss in segmentation tasks and mean square error in super resolution tasks. The training dataset was augmented by applying random rotation (within the range [-15$^\circ$, +15$^\circ$]), random shifting (in both x- and y-directions; within the range of [-5\%, 5\%]), random shearing, random zooming (within the range [0, 0.2]), and random flipping (both horizontally and vertically). The batch size is set to 2 in both training and testing phases. Unless explicitly specified, our BiO-Net is constructed with an encoding depth of 4 and a backward skip connection built at each stage of the network. Our BiO-Net was trained by 300 epochs in all experiments, which were conducted on a single NVIDIA GeForce GTX 1080 GPU with Keras. Given the GPU limitation, we explore the performance improvement to its maximum possible temporal step at \textit{t}=3.
	
	%For semantic segmentation, we use the cross entropy loss as the objective function. Following \cite{graham2018hover}, we train our nuclei segmentation models on MoNuSeg training set only and test our models on MoNuSeg testing set and TNBC dataset. We report dice coefficient (DICE) and Intersection over Union (IoU) when comparing our BiO-Net with others. 
	
	%For super resolution task, we choose mean square error as the reconstruction loss, while Peak Signal to Noise Ratio (PSNR) is reported as the quantitative evaluation metric for comparisons.

	%18273898 UNET++ resnet18 backbone 18.27 M 0.6819, 0.8092, 0.2759 0.4022
	%UNET++ resnet50 backbone 37.70 M 
	
	%						MoNuSeg		     TNBC		#param	model size
	
	%linknet t=3,	0.634 0.774 	0.571 0.716  11.58 M   140 mb
	%linknet t=1,	0.625 0.767 	0.535 0.682 11.54 M   139 mb
	%linknet t=2,  0.621 0.766      0.541  0.690  11.56 M  139 MB
	\begin{table}[t]
		\caption{Comparison of segmentation methods on MoNuSeg testing set and TNBC.}\label{tab1}
		\centering
		\begin{tabular}{{l|c c|c c|c | c}}
			\toprule
			\multicolumn{1}{c}{} & \multicolumn{2}{c}{MoNuSeg} & \multicolumn{2}{c}{TNBC}& 	\multicolumn{2}{c}{}\\
			\hline
			Methods &  IoU & DICE & IoU & DICE &\#params&model size\\
			\hline
			\hline
			U-Net \cite{ronneberger2015u} \textbf{w.} ResNet-18 \cite{he2016deep}&  0.684 & 0.810 & 0.459 & 0.603&15.56 M&62.9 MB\\
			U-Net++ \cite{zhou2018unetpp} \textbf{w.} ResNet-18 \cite{he2016deep} &0.683&0.811&0.526&0.652&18.27 M& 74.0 MB\\
			U-Net++ \cite{zhou2018unetpp} \textbf{w.} ResNet-50 \cite{he2016deep} &0.695&0.818&0.542&0.674&37.70 M& 151.9 MB\\
			Micro-Net \cite{raza2019micro} & 0.696 & 0.819  & 0.544&0.701&14.26 M&57.4 MB\\
			Naylor \textit{et al}. \cite{naylor2018segmentation} & 0.690 & 0.816& 0.482 &0.623 &36.63 M&146.7 MB\\
			M-Net \cite{mehta2017m} &0.686&0.813&0.450&0.569 &0.6 M&2.7 MB\\
			%Hover-Net \cite{graham2018hover} & - & 0.826 & -&0.749\\
			Att U-Net \cite{oktay2018attention} &0.678  & 0.810 &  0.581 & 0.717&33.04 M&133.2 MB\\
			\hline
			R2U-Net, \textit{t}=2 \cite{alom2018nuclei} & 0.678  & 0.807 & 0.532 & 0.650&37.02 M&149.2 MB\\
			R2U-Net, \textit{t}=3 \cite{alom2018nuclei} &0.683  & 0.815& 0.590 & 0.711 &37.02 M&149.2 MB\\
			\hline
			LinkNet \cite{chaurasia2017linknet} & 0.625  & 0.767 & 0.535 & 0.682&11.54 M&139.4 MB\\
			BiO-LinkNet \cite{chaurasia2017linknet}, \textit{t}=2 (Ours) & 0.621  & 0.766 & 0.541 & 0.690&11.54 M&139.4 MB\\ 
			BiO-LinkNet \cite{chaurasia2017linknet}, \textit{t}=3 (Ours) & 0.634  & 0.774 & 0.571 & 0.716&11.54 M&139.4 MB\\ 
			\hline
			BiO-Net, \textit{t}=1 (Ours)& 0.680 & 0.803 & 0.456 & 0.608 &15.0 M& 60.6 MB\\
			BiO-Net, \textit{t}=2 (Ours)& 0.694 & 0.816& 0.548 & 0.693 &15.0 M& 60.6 MB\\
			BiO-Net, \textit{t}=3 (Ours)& 0.700 & 0.821& 0.618 & 0.751 &15.0 M& 60.6 MB\\
			BiO-Net, \textit{t}=3, \texttt{INT} (Ours)& \textbf{0.704}  & \textbf{0.825}& \textbf{0.651} & \textbf{0.780} &15.0 M& 60.6 MB\\
			\bottomrule
		\end{tabular}
	\end{table}
	
	\subsection{Semantic Segmentation}
\begin{figure}[]
		\centering
		\includegraphics[width=\textwidth]{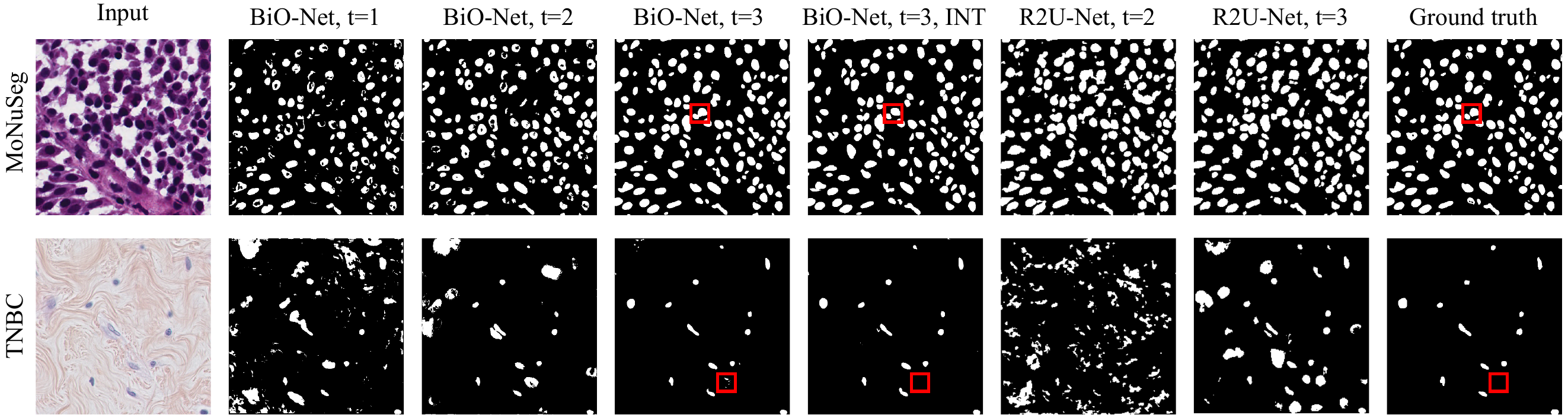}
		\caption{Qualitative comparison between our models and the R2U-Net \cite{alom2018nuclei} on the MoNuSeg testing set and TNBC. Red boundary boxes indicate the effects of integrating features from each iteration at the last stage block.} \label{fig3}
	\end{figure}

	\textbf{Nuclei Segmentation.} In this task, our method is compared to the baseline U-Net \cite{ronneberger2015u} and other state-of-the-art methods \cite{raza2019micro,naylor2018segmentation,zhou2018unetpp,mehta2017m,oktay2018attention,alom2018nuclei}. Following \cite{graham2018hover}, models were trained on the MoNuSeg training set only and evaluated on the MoNuSeg testing set and TNBC dataset. Dice coefficient (DICE) and Intersection over Union (IoU) are evaluated. As shown in Table \ref{tab1}, our results are better than others on the MoNuSeg testing set. Our results on the TNBC dataset are also higher than the others by a large margin, which demonstrates a strong generalization ability. Additionally, compared with other extensions of U-Net, and, our BiO-Net is more memory efficient. Qualitative comparison of our method and the recurrent counterpart R2U-Net \cite{alom2018nuclei} is shown in Fig. \ref{fig3}. It can be seen that our model segments nuclei more accurately as the inference time increases. In our experiments, BiO-Net infers a batch of two predictions in 35, 52, and 70 ms when \textit{t}=1, \textit{t}=2, and \textit{t}=3, respectively. Further evaluation of incorporating our method into another encoder-decoder architecture LinkNet~\cite{chaurasia2017linknet}, which is also shown in the table. Our BiO-LinkNet adds the skipped features element-wisely and, hence, shares the same number of parameters as the vanilla LinkNet.

	\begin{table} [t]
		\caption{Ablative results. The parameters are defined as depicted in Sec. \ref{arch}. IoU(DICE), number of parameters, and, model size are reported.}\label{tab2}
		\centering
		\resizebox{\linewidth}{!}{\begin{tabular}{{c| c c c |c c c | c | c}}
				\toprule
				\multicolumn{1}{c}{} & \multicolumn{3}{c}{MoNuSeg} & \multicolumn{3}{c}{TNBC}& \multicolumn{2}{c}{}\\
				\hline
				&  \textit{t}=1 & \textit{t}=2 & \textit{t}=3 &  \textit{t}=1 & \textit{t}=2 & \textit{t}=3 & \#params & model size \\
				\hline
				\hline
				$\times$1.25 & 0.685(0.813) & 0.698(0.819)& 0.695(0.817)& 0.490(0.637)& 0.557(0.697)&0.623(0.758)&23.5 M & 94.3 MB\\
				$\times$1.0 & 0.680(0.803) & 0.694(0.816) & 0.700(0.821)& 0.456(0.608)& 0.548(0.693)&0.618(0.751)& 15.0 M & 60.6 MB\\
				$\times$0.75 & 0.676(0.800)& 0.678(0.805)& 0.691(0.815)&0.516(0.661)&0.571(0.710) &0.598(0.738)& 8.5 M&34.3 MB\\
				$\times$0.5 & 0.668(0.792)& 0.680(0.806) & 0.691(0.814)& 0.491(0.644) &  0.543(0.679)&0.611(0.742) & 3.8 M& 15.8 MB\\
				$\times$0.25 & 0.667(0.791)& 0.678(0.804)& 0.677(0.804)& 0.524(0.674)&0.535(0.678) &0.575(0.710) & 0.9 M& 4.0 MB   \\
				\hline
				$w$=3 & 0.680(0.803) & 0.694(0.817) & 0.688(0.814) &0.456(0.608) &0.510(0.656)& 0.620(0.757) &15.0 M & 60.3 MB\\
				$w$=2& 0.680(0.803) & 0.672(0.801)& 0.686(0.813) &  0.456(0.608) &0.527(0.679)& 0.601(0.742) & 14.9 M & 60.1 MB\\
				\hline
				\texttt{INT}& 0.680(0.803) & 0.689(0.812)& \textbf{0.704(0.825)}& 0.456(0.608) & 0.588(0.728)&\textbf{0.651(0.780)}&15.0 M &60.6 MB\\
				\hline
				$l$=3& 0.681(0.806) & 0.679(0.805)& 0.689(0.812) & 0.613(0.742) & 0.594(0.733)& 0.615(0.741) &3.8 M& 15.4 MB \\
				$l$=2& 0.690(0.810) & 0.695(0.817) & 0.697(0.818)& 0.596(0.734)&0.647(0.775)&0.596(0.735)& 0.9 M& 4.0 MB\\
				\bottomrule
		\end{tabular}}
	\end{table}
	%Single tail paired t-test was performed to evaluate the significance of the results, which yields a p-value less than 0.01. 
	\noindent Table \ref{tab2} demonstrates our ablation study by varying the setups as defined in Fig. \ref{fig2}. The results show that recursing through the encoder and the decoder with the proposed bi-directional skip connections improves network performances generally. Integrating decoded features from all inference recurrences yields state-of-the-art performances in both datasets. Furthermore, we find that when there are insufficient parameters in the network, increasing inference recurrence has little improvement or even makes the results worse. It is also interesting to observe that when constructing BiO-Net with shallower encoding depth, our models perform better on the two datasets than those with deeper encoding depth. \\
	\noindent
	\textbf{EM Membrane Segmentation.} We further evaluated our method by segmenting Mouse Piriform Cortex EM images \cite{lee2015recursive}, where the models are trained on stack1 and stack2, and validated on stack4 and stack3. The results were evaluated by Rand F-score \cite{arganda2015crowdsourcing}. As shown in Table \ref{tab3}, our method demonstrates better segmentation results with the proposed bi-directional O-shaped skip connections.
	
	%\footnote{The script in \url{http://imagej.net/Segmentation_evaluation_after_border_thinning_-_Script}}
	
	%stack 3
	%rand score 0.9552 t=2
	%rand score 0.9584 t=3
	%rand score 0.9415  t=1
	%rand score 0.9298 unet
	%rand score 0.9203 Xnet
	
	%stack 4
	%rand score 0.8712 t=2
	%rand score 0.8872 t=3
	%rand score 0.8279  t=1
	%rand score 0.8338  att unet
	%rand score 0.6215 unet
	%rand score 0.5247 Xnet
	
	%rand score 0.9271 unet
	%rand score 0.9334 Xnet
	
	\begin{table} [t]
		\caption{Comparison of different U-Net variants in EM membrane segmentation.}\label{tab3}
		\centering
		\begin{tabular}{{c | c  c c || c |c c c}}
			\toprule
			Variants & stack3& stack4 & \#params & Ours& stack3& stack4&\#params\\
			\hline
			\hline
			U-Net & 0.939&0.821&15.56M& BiO-Net, t=1& 0.941&0.827 &15.0M\\
			Att U-Net & 0.937&0.833&33.04M  &BiO-Net, t=2& 0.955 &0.871 &15.0M \\
			U-Net++ &0.940&0.844&18.27M&BiO-Net, t=3&\textbf{0.958}&\textbf{0.887}&15.0M \\
			\bottomrule
		\end{tabular}
	\end{table}
	
	%We demonstrate the qualitative results of the setup with the best performance in Fig. \ref{fig3}. In this figure, it is clear that our model segments nuclei better as the inference time increases. We highlight the comparison details in this figure, which is obvious that by integrating features from all inference recurrences, less noise will be included, whereas risks at omitting non-obvious details.

	%Based on the above observation, we further experiment to construct an BiO-Net model with $\times$0.25 features maps and encoding depth 2 for a compact and powerful design. The compact model yields the results 
	
	%\begin{table} 
	%	\caption{Comparision of different nuclei segmentation methods over MoNuSeg testing set and TNBC dataset.}\label{tab2}
	%	\centering
	%	\begin{tabular}{{c c c c c c c c c c c}}
	
	%	\toprule
	%		 &  $\times$1.25 & $\times$0.75 & $\times$0.5 & $\times$0.25 & w=3& w=2 & INT &l=3 & l=2 & l=1\\
	%		\hline
	%		t = 1 &  & &  & &  & &  & & &\\
	%		t = 2 &  & &  & &  & &  & & &\\
	%		t = 3 &  & &  & &  & &  & & &\\
	%		\bottomrule
	%	\end{tabular}
	%\end{table}
	
	%\paragraph{Reduce Parameter Size.}
	%\paragraph{Reduce Backward Connections.}
	%\paragraph{Integrate inference.}
	%\paragraph{Reduce encoding depth.}

	\subsection{Super Resolution}  
\begin{figure}[h]
		\centering
		\includegraphics[width=\textwidth]{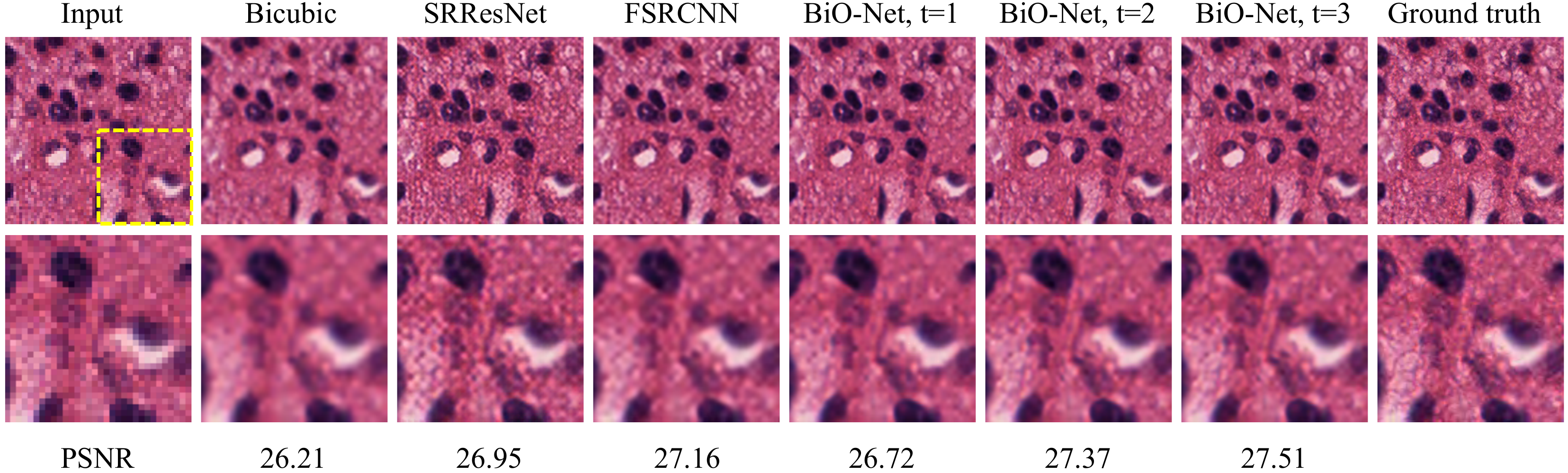}
		\caption{Comparison of different methods in super resolution task. PSNR scores of the methods over the entire testing set are reported as well. The second row is projected from the yellow boundary box to have a closer view on the super resolution achieved in higher resolution.} \label{fig4}
	\end{figure}

	In addition to segmentation tasks, we are also interested in experimenting with our BiO-Net on a significantly different task: image super resolution, which has been studied actively \cite{chen2018efficient,sui2019isotropic}. In the super resolution task, low-resolution (downsampled) images are used as inputs to train the networks toward their original high-resolution ground truth, which can assist medical imaging analysis by recovering missing details and generating high-resolution histopathology images based on the low-resolution ones. Two state-of-the-art methods, FSRCNN \cite{dong2016accelerating} and SRResNet \cite{ledig2017photo}, are adopted to compare with our BiO-Net. The qualitative results along with the Peak Signal to Noise Ratio (PSNR) score over the entire testing set are shown in Fig. \ref{fig4}. It can be seen that, our method outperforms the state-of-the-art methods by a safe margin, which validates the feasibility of applying our BiO-Net on different visual tasks.
	
	\section{Conclusion}
	
	In this paper, we introduced a novel recurrent variant of U-Net, named BiO-Net. BiO-Net is a compact substitute of U-Net with better performance and no extra trainable parameters, which utilizes paired forward and backward skip connections to compose a complex relation between the encoder and decoder. The model can be recursed to reuse the parameters during training and inference. Extensive experiments on semantic segmentation and super-resolution tasks indicate the effectiveness of our proposed model, which outperforms the U-Net and its extension methods without introducing auxiliary parameters.

%	{\small
%		\bibliographystyle{splncs04}
%		\bibliography{ref}

\begin{thebibliography}{10}
		\bibliographystyle{splncs04}
		\providecommand{\url}[1]{\texttt{#1}}
		\providecommand{\urlprefix}{URL }
		\providecommand{\doi}[1]{https://doi.org/#1}
		
		\bibitem{alom2018nuclei}
		Alom, M.Z., Yakopcic, C., Taha, T.M., Asari, V.K.: Nuclei segmentation with
		recurrent residual convolutional neural networks based u-net (r2u-net). In:
		IEEE National Aerospace and Electronics Conference. pp. 228--233. IEEE (2018)
		
		\bibitem{arganda2015crowdsourcing}
		Arganda-Carreras, I., Turaga, S.C., Berger, D.R., Cire{\c{s}}an, D., Giusti,
		A., Gambardella, L.M., Schmidhuber, J., Laptev, D., Dwivedi, S., Buhmann,
		J.M., et~al.: Crowdsourcing the creation of image segmentation algorithms for
		connectomics. Frontiers in Neuroanatomy  \textbf{9}, ~142 (2015)
		
		\bibitem{chaurasia2017linknet}
		Chaurasia, A., Culurciello, E.: Linknet: Exploiting encoder representations for
		efficient semantic segmentation. In: IEEE Visual Communications and Image
		Processing (VCIP). pp.~1--4. IEEE (2017)
		
		\bibitem{chen2018efficient}
		Chen, Y., Shi, F., Christodoulou, A.G., Xie, Y., Zhou, Z., Li, D.: Efficient
		and accurate mri super-resolution using a generative adversarial network and
		3d multi-level densely connected network. In: International Conference on
		Medical Image Computing and Computer-Assisted Intervention (MICCAI). pp.
		91--99. Springer (2018)
		
		\bibitem{dong2016accelerating}
		Dong, C., Loy, C.C., Tang, X.: Accelerating the super-resolution convolutional
		neural network. In: European Conference on Computer Vision (ECCV). pp.
		391--407. Springer (2016)
		
		\bibitem{graham2018hover}
		Graham, S., Vu, Q.D., Raza, S.E.A., Azam, A., Tsang, Y.W., Kwak, J.T., Rajpoot,
		N.: Hover-net: Simultaneous segmentation and classification of nuclei in
		multi-tissue histology images. Medical Image Analysis (MIA)  \textbf{58},
		101563 (2019)
		
		\bibitem{guo2019dynamic}
		Guo, Q., Yu, Z., Wu, Y., Liang, D., Qin, H., Yan, J.: Dynamic recursive neural
		network. In: Proceedings of the IEEE Conference on Computer Vision and
		Pattern Recognition (CVPR). pp. 5147--5156 (2019)
		
		\bibitem{han2018image}
		Han, W., Chang, S., Liu, D., Yu, M., Witbrock, M., Huang, T.S.: Image
		super-resolution via dual-state recurrent networks. In: Proceedings of the
		IEEE Conference on Computer Vision and Pattern Recognition (CVPR). pp.
		1654--1663 (2018)
		
		\bibitem{he2016deep}
		He, K., Zhang, X., Ren, S., Sun, J.: Deep residual learning for image
		recognition. In: Proceedings of the IEEE Conference on Computer Vision and
		Pattern Recognition (CVPR). pp. 770--778 (2016)
		
		\bibitem{hou2019robust}
		Hou, L., Agarwal, A., Samaras, D., Kurc, T.M., Gupta, R.R., Saltz, J.H.: Robust
		histopathology image analysis: To label or to synthesize? In: Proceedings of
		the IEEE Conference on Computer Vision and Pattern Recognition (CVPR). pp.
		8533--8542 (2019)
		
		\bibitem{ioffe2015batch}
		Ioffe, S., Szegedy, C.: Batch normalization: Accelerating deep network training
		by reducing internal covariate shift. In: International Conference on Machine
		Learning (ICML). pp. 448--456 (2015)
		
		\bibitem{kingma:adam}
		Kingma, D.P., Ba, J.: Adam: A method for stochastic optimization. In:
		International Conference on Learning Representations (ICLR) (2015)
		
		\bibitem{kumar2017dataset}
		Kumar, N., Verma, R., Sharma, S., Bhargava, S., Vahadane, A., Sethi, A.: A
		dataset and a technique for generalized nuclear segmentation for
		computational pathology. IEEE Transactions on Medical Imaging (TMI)
		\textbf{36}(7),  1550--1560 (2017)
		
		\bibitem{ledig2017photo}
		Ledig, C., Theis, L., Husz{\'a}r, F., Caballero, J., Cunningham, A., Acosta,
		A., Aitken, A., Tejani, A., Totz, J., Wang, Z., et~al.: Photo-realistic
		single image super-resolution using a generative adversarial network. In:
		Proceedings of the IEEE Conference on Computer Vision and Pattern Recognition
		(CVPR). pp. 4681--4690 (2017)
		
		\bibitem{lee2015recursive}
		Lee, K., Zlateski, A., Ashwin, V., Seung, H.S.: Recursive training of 2d-3d
		convolutional networks for neuronal boundary prediction. In: Advances in
		Neural Information Processing Systems (NeurIPS). pp. 3573--3581 (2015)
		
		\bibitem{mehta2017m}
		Mehta, R., Sivaswamy, J.: M-net: A convolutional neural network for deep brain
		structure segmentation. In: 14th International Symposium on Biomedical
		Imaging (ISBI). pp. 437--440. IEEE (2017)
		
		\bibitem{milletari2016v}
		Milletari, F., Navab, N., Ahmadi, S.A.: V-net: Fully convolutional neural
		networks for volumetric medical image segmentation. In: 4th International
		Conference on 3D Vision (3DV). pp. 565--571. IEEE (2016)
		
		\bibitem{nair2010rectified}
		Nair, V., Hinton, G.E.: Rectified linear units improve restricted boltzmann
		machines. In: Proceedings of the 27th International Conference on Machine
		Learning (ICML). pp. 807--814 (2010)
		
		\bibitem{naylor2018segmentation}
		Naylor, P., La{\'e}, M., Reyal, F., Walter, T.: Segmentation of nuclei in
		histopathology images by deep regression of the distance map. IEEE
		Transactions on Medical Imaging (TMI)  \textbf{38}(2),  448--459 (2018)
		
		\bibitem{oktay2018attention}
		Oktay, O., Schlemper, J., Folgoc, L.L., Lee, M., Heinrich, M., Misawa, K.,
		Mori, K., McDonagh, S., Hammerla, N.Y., Kainz, B., et~al.: Attention u-net:
		Learning where to look for the pancreas. 1st Conference on Medical Imaging
		with Deep Learning (MIDL)  (2018)
		
		\bibitem{raza2019micro}
		Raza, S.E.A., Cheung, L., Shaban, M., Graham, S., Epstein, D., Pelengaris, S.,
		Khan, M., Rajpoot, N.M.: Micro-net: A unified model for segmentation of
		various objects in microscopy images. Medical Image Analysis (MIA)
		\textbf{52},  160--173 (2019)
		
		%\bibitem{romera2016recurrent}
		%Romera-Paredes, B., Torr, P.H.S.: Recurrent instance segmentation. In: European
		%Conference on Computer Vision (ECCV). pp. 312--329. Springer (2016)
		
		\bibitem{ronneberger2015u}
		Ronneberger, O., Fischer, P., Brox, T.: U-net: Convolutional networks for
		biomedical image segmentation. In: International Conference on Medical Image
		Computing and Computer-Assisted Intervention (MICCAI). pp. 234--241. Springer
		(2015)
		
		\bibitem{sui2019isotropic}
		Sui, Y., Afacan, O., Gholipour, A., Warfield, S.K.: Isotropic mri
		super-resolution reconstruction with multi-scale gradient field prior. In:
		International Conference on Medical Image Computing and Computer-Assisted
		Intervention (MICCAI). pp. 3--11. Springer (2019)
		
		\bibitem{wang2019recurrent}
		Wang, W., Yu, K., Hugonot, J., Fua, P., Salzmann, M.: Recurrent u-net for
		resource-constrained segmentation. In: The IEEE International Conference on
		Computer Vision (ICCV) (2019)
		
		\bibitem{xia2017w}
		Xia, X., Kulis, B.: W-net: A deep model for fully unsupervised image
		segmentation. arXiv preprint arXiv:1711.08506  (2017)
		
		\bibitem{zhang2018ms}
		Zhang, C., Song, Y., Liu, S., Lill, S., Wang, C., Tang, Z., You, Y., Gao, Y.,
		Klistorner, A., Barnett, M., et~al.: Ms-gan: Gan-based semantic segmentation
		of multiple sclerosis lesions in brain magnetic resonance imaging. In: 2018
		Digital Image Computing: Techniques and Applications (DICTA). pp.~1--8. IEEE
		(2018)
		
		\bibitem{zhang2018whole}
		Zhang, C., Song, Y., Zhang, D., Liu, S., Chen, M., Cai, W.: Whole slide image
		classification via iterative patch labelling. In: 25th IEEE International
		Conference on Image Processing (ICIP). pp. 1408--1412. IEEE (2018)
		
		\bibitem{zhou2018unetpp}
		Zhou, Z., Siddiquee, M.M.R., Tajbakhsh, N., Liang, J.: Unet++: A nested u-net
		architecture for medical image segmentation. In: Deep Learning in Medical
		Image Analysis and Multimodal Learning for Clinical Decision Support (DLMIA),
		pp. 3--11. Springer (2018)
		
	\end{thebibliography}
	%}

\end{document}